\documentclass[twoside,11pt]{article}
\usepackage[preprint]{jmlr2e}

\usepackage[utf8]{inputenc} 
\usepackage[T1]{fontenc}    
\usepackage{url}            
\usepackage{booktabs}       
\usepackage{amsfonts}       
\usepackage{nicefrac}       
\usepackage{microtype}      
\usepackage{xcolor}         
\usepackage{pifont}         
\usepackage{breakcites}

\usepackage{array}
\newcommand{\PreserveBackslash}[1]{\let\temp=\\#1\let\\=\temp}

\newcolumntype{C}[1]{>{\PreserveBackslash\centering}p{#1}}
\newcolumntype{R}[1]{>{\PreserveBackslash\raggedleft}p{#1}}
\newcolumntype{L}[1]{>{\PreserveBackslash\raggedright}p{#1}}

\hypersetup{
    colorlinks=true,
    allcolors=brown
}
\usepackage{physics} 
\usepackage[linesnumbered,ruled,vlined]{algorithm2e}
\usepackage{xcolor}
\usepackage{amsmath}
\usepackage{amssymb}
\usepackage{numprint}
\usepackage{dsfont}
\usepackage{bm}
\usepackage{bbm}
\usepackage{float}
\usepackage{sectsty}
\usepackage{graphicx} 
\usepackage{subcaption}
\usepackage{enumitem}
\usepackage{authblk}
\usepackage{multido}
\usepackage{listings}
\usepackage{xcolor}

\definecolor{codegreen}{rgb}{0,0.6,0}
\definecolor{codegray}{rgb}{0.5,0.5,0.5}
\definecolor{codepurple}{rgb}{0.58,0,0.82}
\definecolor{backcolour}{rgb}{0.95,0.95,0.92}

\lstdefinestyle{mystyle}{
    commentstyle=\color{codegreen},
    keywordstyle=\color{magenta},
    numberstyle=\tiny\color{codegray},
    stringstyle=\color{codepurple},
    basicstyle=\ttfamily\footnotesize,
    breakatwhitespace=false,         
    breaklines=true,                 
    captionpos=b,                    
    keepspaces=true,                 
    numbers=left,                    
    numbersep=5pt,                  
    showspaces=false,                
    showstringspaces=false,
    showtabs=false,                  
    tabsize=2
}

\newcommand{\cmark}{\ding{51}}%

\def\x{{\bf x}}

\def\0{{\bf 0}}
\def\1{{\bf 1}}

\makeatletter
\def\@fnsymbol#1{\ensuremath{\ifcase#1\or \dagger\or \ddagger\or
   \mathsection\or \mathparagraph\or \|\or **\or \dagger\dagger
   \or \ddagger\ddagger \else\@ctrerr\fi}}
\makeatother
    
\ShortHeadings{OmniXAI: A Library for Explainable AI}{Yang et al.}
\firstpageno{1}

\title{OmniXAI: A Library for Explainable AI}

\author[1,*]{Wenzhuo Yang}
\author[1]{Hung Le}
\author[2]{Tanmay Laud \thanks{Tanmay made contribution when he was an intern in Salesforce.}}
\author[1]{Silvio Savarese}
\author[1,*]{Steven C.H. Hoi}
\affil{Salesforce Research}
\affil[2]{Salesforce Search}
\affil[1,*]{Corresponding Authors: \texttt{\{wenzhuo.yang,shoi\}@salesforce.com}}

\begin{document}
\maketitle

\begin{abstract}
We introduce OmniXAI\footnote{https://github.com/salesforce/OmniXAI} (short for Omni eXplainable AI), an open-source Python library of eXplainable AI (XAI), which offers omni-way explainable AI capabilities and various interpretable machine learning techniques to address the pain points of understanding and interpreting the decisions made by machine learning (ML) in practice. OmniXAI aims to be a one-stop comprehensive library that makes explainable AI easy for data scientists, ML researchers and practitioners who need explanation for various types of data, models and explanation methods at different stages of a ML process (data exploration, feature engineering, model development, evaluation, and decision-making, etc). In particular, our library includes a rich family of explanation methods integrated in a unified interface, which supports multiple data types (tabular data, images, texts, time-series), multiple types of ML models (traditional ML in Scikit-learn and deep learning models in PyTorch/TensorFlow), and a range of diverse explanation methods including ``model-specific'' and ``model-agnostic'' ones (such as feature-attribution explanation, counterfactual explanation, gradient-based explanation, etc). For practitioners, the library provides an easy-to-use unified interface to generate the explanations for their applications by only writing a few lines of codes, and also a GUI dashboard for visualization of different explanations for more insights about decisions. In this technical report, we present OmniXAI's design principles, system architectures, and major functionalities, and also demonstrate several example use cases across different types of data, tasks, and models.
\end{abstract}


\section{Introduction}

With the rapidly growing adoption of machine learning (ML) models in real-world applications, the algorithmic decisions of ML models can potentially have a significant societal impact on building trustworthy AI systems for real-world applications, especially for some application domains such as healthcare, education, and finance, etc. Many ML models, especially deep learning models, work as black-box models that lack explainability and thus inhibit their adoption in some critical applications and hamper the trust in machine learning/AI systems. To address these challenges, eXplainable AI (XAI) is an emerging field in machine learning and AI, aiming to explain how those black-box decisions of AI systems are made. Such explanations can improve the transparency, persuasiveness, and trustworthiness of AI systems and help AI developers and practitioners to debug and improve model performance ~\citep{doshivelez2017rigorous,Kusner2017,Lipton2016,JMLR:v22:20-1473,JMLR:v22:21-0017}. 

To this end, we develop OmniXAI (short for Omni eXplainable AI), an open-source library for explainable AI to provide omni-way explainability for a variety of machine learning models. 
\begin{figure}[!ht]
\center
\includegraphics[width=1.0\linewidth]{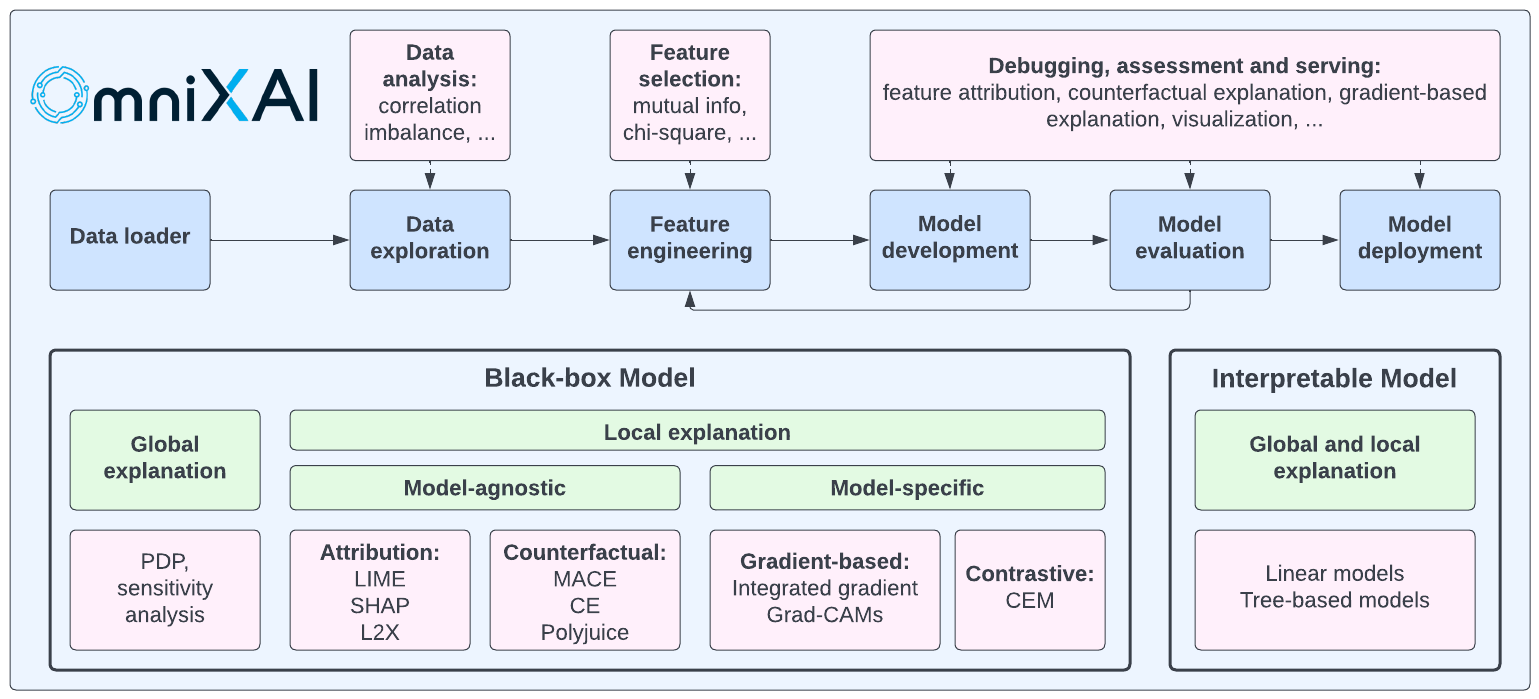}
\caption{OmniXAI offers explainablity for different stages in a standard ML workflow.}
\label{fig:ml_pipeline}
\end{figure}

As shown in Figure~\ref{fig:ml_pipeline}, OmniXAI offers a one-stop solution for analyzing different stages in a standard ML workflow in real-world applications, and provides a comprehensive family of explanation methods, including high-quality implementations of various model-agnostic and model-specific explanation methods. In data analysis and exploration, OmniXAI can analyze feature correlations and data imbalance issues, helping developers quickly remove redundant features and discover potential bias issues. In feature engineering, OmniXAI can identify important features by analyzing relationships between features and targets, helping them understand data properties and do feature preprocessing. In model training and evaluation, OmniXAI provides various explanations, e.g., feature-attribution explanation, counterfactual explanation, gradient-based explanation, to comprehensively analyze the behavior of a model designed for tabular, vision, NLP, or time-series tasks. It helps to debug the model when wrong predictions occur or the model behavior is not as expected, and gathers more valuable information for improving feature engineering. After the model is deployed, OmniXAI opens ``black boxes'' and generates explanations for each model decision so that domain experts can analyze the decisions and explanations to determine which actions to take.

Compared with other existing explanation libraries, e.g., IBM’s AIX360~\citep{aix360-sept-2019}, Microsoft’s InterpretML \citep{nori2019interpretml}, DALEX \citep{JMLR:v22:20-1473} and Alibi \citep{JMLR:v22:21-0017}, our library has a comprehensive list of XAI capabilities and unique features including:
\begin{itemize}
    \item \textbf{Support data analysis/exploration}: we support feature analysis and selection, e.g., feature correlations, data imbalance issues, selecting important features.
    \item \textbf{Support popular machine learning frameworks}: we support the most popular machine learning frameworks or models, e.g., PyTorch, Tensorflow, scikit-learn and customized black-box models.
    \item \textbf{Support most popular explanation methods}: Many explanation methods, e.g., feature-attribution explanation, counterfactual explanation, partial dependence plots, etc., are included for analyzing different aspects of a machine learning model.
    \item \textbf{Support explanations for tabular ranking tasks:} We provide model-agnostic feature-attribution explanations for black-box learning-to-rank models. To the best of our knowledge, ours is the only library with this utility.
    \item \textbf{Support counterfactual explanations}: An efficient counterfactual explanation algorithm called MACE (model-agnostic counterfactual explanation) \citep{mace} designed for tabular and time-series data is included in the library. For text data, we include a method based on Polyjuice \cite{wu-etal-2021-polyjuice} for text classification and question-answering tasks.
    \item \textbf{Support gradient-based explanations}: Integrated-gradient \citep{10.5555/3305890.3306024}, Grad-CAM \citep{Grad-CAM2019} and Grad-CAM's variants \citep{8354201} are supported.
    \item \textbf{Support image, text, and time-series data}: OmniXAI provides multiple explanation methods for a diverse set of tasks,
    e.g., Grad-CAM for vision tasks, integrated-gradient for NLP tasks, counterfactual for time-series tasks.
    \item \textbf{Support feature visualization}: OmniXAI supports feature visualization (an optimization-based method), and also provides ability to visualize feature maps for each layer.
    \item \textbf{Easy-to-use}: Users only need to write a few lines of code to generate various kinds of explanations, and OmniXAI supports Jupyter Notebook environments.
    \item \textbf{Easy-to-extend}: Developers can add new explanation algorithms easily by implementing a single class inherited from the explainer base class.
    \item \textbf{A GUI dashboard}: OmniXAI provides a visualization tool for users to examine the generated explanations and compare interpretability algorithms.
\end{itemize}

Table~\ref{table_1} shows the supported interpretability algorithms in OmniXAI\footnote{SHAP accepts black-box models for tabular data, PyTorch/Tensorflow models for image data, transformer models for text data. Counterfactual explanation accepts black-box models for tabular data and PyTorch/Tensorflow models for image data.}, and Table~\ref{table_2} shows the comparison between OmniXAI and other existing XAI toolkits/libraries in literature. Our library supports all of these methods with a unified interface.
\begin{table}[!ht]
\caption{The supported interpretability algorithms in OmniXAI}\vspace{-0.2in}
\label{table_1}
\center
\begin{small}
\begin{tabular}{|c|c|c|c|c|c|c|c|}
  \hline
  Method & Model Type & Exp Type & EDA & Tabular & Image & Text & TS \\ \hline
  Feature analysis & NA & Global & \cmark & & & &\\ \hline
  Feature selection   & NA & Global & \cmark & & & & \\ \hline
  Partial dependence plot & Black box & Global & & \cmark & & & \\ \hline
  Accumulated local effects & Black box & Global & & \cmark & & & \\ \hline
  Sensitivity analysis & Black box & Global & & \cmark & & & \\ \hline
  Feature visualization & Torch or TF & Global & & & \cmark & & \\ \hline
  LIME & Black box & Local & & \cmark & \cmark & \cmark & \\ \hline
  SHAP & Black box* & Local & & \cmark & \cmark & \cmark & \cmark \\ \hline
  Integrated gradient & Torch or TF & Local & & \cmark & \cmark & \cmark & \\ \hline
  Counterfactual & Black box* & Local & & \cmark & \cmark & \cmark & \cmark \\ \hline
  Contrastive explanation & Torch or TF & Local & & & \cmark & & \\ \hline
  Grad-CAM & Torch or TF & Local & & & \cmark & & \\ \hline
  Learning to explain & Black box & Local & & \cmark & \cmark & \cmark & \\ \hline
  Linear models & Linear models & Global/local & & \cmark & & & \\ \hline
  Tree models & Tree models & Global/local & & \cmark & & & \\ \hline
  Feature maps & Torch or TF & Local & & & \cmark & & \\ \hline
\end{tabular}
\end{small}
\end{table}
\begin{table}[!ht]
\caption{Comparison between OmniXAI and other XAI libraries.}\vspace{-0.2in}
\label{table_2}
\center
\begin{small}
\begin{tabular}{|c|c|c|c|c|c|c|c|}
  \hline
  Data Type & Method & OmniXAI & InterpretML & AIX360 & Eli5 & Captum & Alibi \\ \hline
  Tabular & LIME                 & \cmark & \cmark & \cmark & & \cmark & \\ \hline
          & SHAP                 & \cmark & \cmark & \cmark & & \cmark & \cmark \\ \hline
          & PDP                  & \cmark & \cmark & & & & \\ \hline
          & ALE                  & \cmark &  & & & & \cmark \\ \hline
          & Sensitivity          & \cmark & \cmark & & & & \\ \hline
          & Integrated gradient  & \cmark & & & & \cmark & \cmark \\ \hline
          & Counterfactual       & \cmark & & & & & \cmark \\ \hline
          & Linear models        & \cmark & \cmark & \cmark & \cmark & & \cmark \\ \hline
          & Tree models          & \cmark & \cmark & \cmark & \cmark & & \cmark \\ \hline
          & L2X                  & \cmark & & & & & \\ \hline
  Image   & LIME                 & \cmark & & & & \cmark & \\ \hline
          & SHAP                 & \cmark & & & & \cmark & \\ \hline
          & Integrated gradient  & \cmark & & & & \cmark & \cmark \\ \hline
          & Grad-CAM             & \cmark & & & \cmark & \cmark & \\ \hline
          & CEM                  & \cmark & & \cmark & & & \cmark \\ \hline
          & Counterfactual       & \cmark & & & & & \cmark \\ \hline
          & L2X                  & \cmark & & & & & \\ \hline
          & Feature visualization & \cmark & & & & & \\ \hline
  Text    & LIME                 & \cmark & & & \cmark & \cmark & \\ \hline
          & SHAP                 & \cmark & & & & \cmark & \\ \hline
          & Integrated gradient  & \cmark & & & & \cmark & \cmark \\ \hline
          & Counterfactual       & \cmark & & & & & \\ \hline
          & L2X                  & \cmark & & & & & \\ \hline
Timeseries & SHAP                & \cmark & & & & & \\ \hline
          & Counterfactual      & \cmark & & & & & \\ \hline
\end{tabular}
\end{small}
\end{table}

The library includes a family of popular model-agnostic explanations methods (such as LIME~\citep{Ribeiro2016}, SHAP~\citep{NIPS2017_7062}, L2X~\citep{chen2018learning}) and model-specific ones (e.g., integrated-gradient (IG)~\citep{10.5555/3305890.3306024}), which generate local explanations and can support multiple data types and tasks. 

For tabular data, OmniXAI includes two methods for global explanations, i.e., partial dependence plot (PDP)~\citep{hastie01statisticallearning} and Morris sensitivity analysis~\citep{10.2307/1269043} for analyzing how each feature affects model outcomes, and two counterfactual explanation methods, i.e.,  CE~\citep{wachter2018a} which can only handle continuous-valued features, and MACE~\citep{mace} which is a model-agnostic method that can handle both continuous-valued and categorical features. 

For image data, OmniXAI also supports Grad-CAM~\citep{Grad-CAM2019}, Grad-CAM++ \citep{8354201} for explaining deep learning models on image domains, the contrastive explanation method (CEM) that finds pertinent negatives and pertinent positives for explanations, and the counterfactual explanation method CE~\citep{wachter2018a}. Note that both CEM and CE only support the classification tasks.

For text data, in addition to the generic explanation methods (such as LIME, SHAP, and IG), OmniXAI also supports a counterfactual explanation method which is based on the pretrained model ``Polyjuice''~\citep{wu-etal-2021-polyjuice} for both classification and QA tasks. 

Finally, for time-series data, OmniXAI supports explanation techniques for both time-series anomaly detection and time-series forecasting. Specifically, it provides a SHAP-based method and an optimization-based counterfactual explanation method MACE for both time-series anomaly detection and forecasting.  

\section{Library Design}

\begin{figure}[t] 
\center
\includegraphics[width=1.0\linewidth]{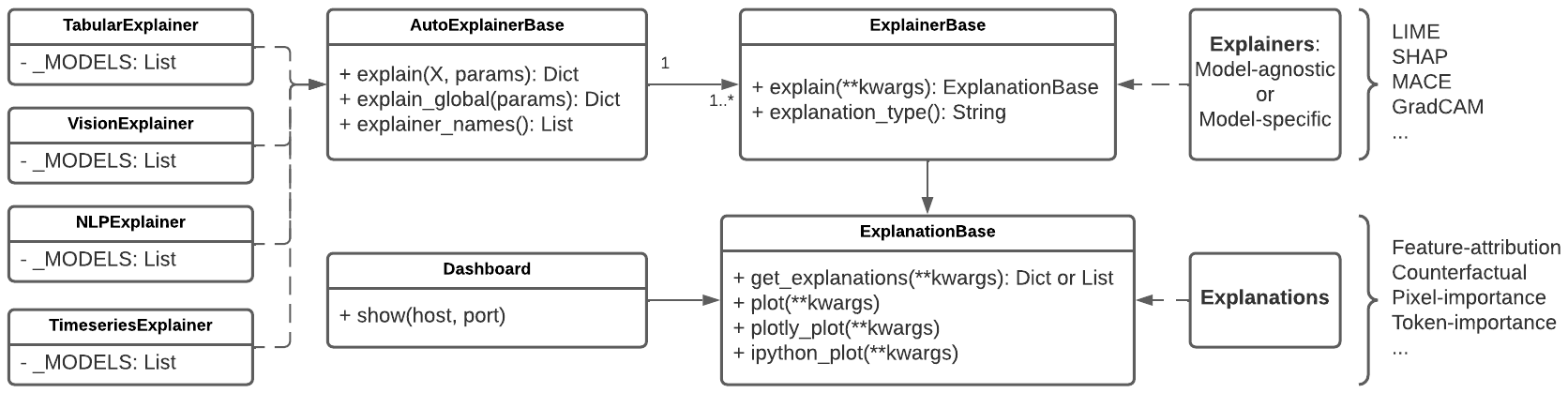}
\caption{The main architecture of OmniXAI.}
\label{fig:class}
\end{figure}

Our key design principle is to provide a simple but unified interface allowing users to apply multiple explanation methods and visualize the corresponding generated explanations at the same time. This aims to make our library easy-to-use (generating explanations by writing a few lines of codes), easy-to-extend (adding new explanation methods easily without affecting the library framework), and easy-to-compare (visualizing the explanation results to compare multiple explanation methods). The library has five key components:
\begin{itemize}
    \item \textbf{Data classes} -- ``omnixai.data'': This package contains the data classes for representing tabular data, image data, text data, and time-series data. For example, the explainers for tabular data use an instance of ``omnixai.data.tabular'' as one of their inputs. The library provides simple constructors for creating instances of these classes from numpy arrays, pandas dataframes, pillow images, or strings.
    \item \textbf{Preprocessing modules} -- ``omnixai.preprocessing'': This package contains various pre-processing functions for different data types. For example, it provides 1) one-hot encoding and ordinal encoding for categorical features, 2) KBins, standard normalization, min-max normalization, rescaling, NaN-filling for continuous-valued features, 3) rescaling, normalization, resizing for image data, and 4) the TF-IDF transformation and token-to-id transformation for text data.
    \item \textbf{Explanation methods} -- ``omnixai.explainers'': This is the main package in the library, which contains all the supported explainers. The explainers are categorized into four groups: 1) ``omnixai.explainers.data'' for data exploration/analysis, including feature correlation analysis, feature selection, etc., 2) ``omnixai.explainers.tabular'' for tabular data, e.g., global explanations such as PDP~\citep{hastie01statisticallearning}, local explanations such as LIME~\citep{Ribeiro2016}, SHAP~\citep{NIPS2017_7062}, MACE~\citep{mace}, 3) ``omnixai.explainers.vision'' for vision tasks, e.g., integrated-gradient \citep{10.5555/3305890.3306024},
    Grad-CAM \citep{Grad-CAM2019}, contrastive explanation \citep{dhurandhar2018explanations}, counterfactual explanation \citep{wachter2018a}, and 4) ``omnixai.explainers.nlp'' for NLP tasks, e.g., LIME, integrated-gradient (IG). 
    
    For each group, the explainers are further categorized into ``model-agnostic'', ``model-specific'' and ``counterfactual''. A ``model-agnostic'' explainer can handle black-box ML models, i.e., only requiring a prediction function without knowing model details. A ``model-specific'' explainer requires some information of ML models, e.g., whether the model is differentiable, whether the model is a linear model or a tree-based model. ``Counterfactual'' is a special group for counterfactual explanation methods.
    \item \textbf{Explanation results} -- ``omnixai.explanations'': This package contains the classes for explanation results. For example, the class ``FeatureImportance'' is used for storing feature-importance/attribution explanations. All of these classes provide plotting functions for visualization, e.g., ``plot'' using ``Matplotlib'', ``plotly\_plot'' using ``Plotly Dash'' and ``ipython\_plot'' for Jupyter Notebook.
    \item \textbf{Visualization tools} -- ``omnixai.visualization'': This package contains a dashboard for visualization implemented using Plotly Dash. The dashboard supports both global explanations and local explanations. It provides a convenient way to compare and analyze multiple explanation results.
\end{itemize} 
Figure \ref{fig:class} demonstrates the main architecture of the library.
The package ``omnixai.explainers'' contains four special explainers, namely ``TabularExplainer'', ``VisionExplainer'', ``NLPExplainer'' and ``TimeseriesExplainer'', inherited from ``AutoExplainerBase'' acting as the factories of the supported explainers. They provide a unified API for generating explanations with multiple explainers.
To initialize them, one only needs to specify the following parameters:
\begin{itemize}
    \item \textbf{The names of the explainers to apply}: e.g., ``shap'' for SHAP, ``pdp'' for PDP, ``gradcam'' for Grad-CAM.
    \item \textbf{The machine learning model to explain}: e.g., a scikit-learn model, a tensorflow model, a pytorch model, or a black-box prediction function.
    \item \textbf{The pre-processing function}: e.g., converting raw input features into the model inputs, and feature processing.
    \item \textbf{The post-processing function (optional)}: e.g., converting model outputs into class probabilities for classification tasks if the outputs are logits.
\end{itemize}
The class ``AutoExplainerBase'' will automatically check whether a chosen explainer is model-agnostic or model-specific and whether it generates local explanations or global explanations, and then determine how to call the method ``explain'' in the class ``ExplainerBase'' correctly to generate explanations. The generated explanations are stored in the explanation classes derived from ``ExplanationBase'' which has one method for extracting raw explanation results and three methods for visualization. One can call ``plot'', ``plotly\_plot'' or ``ipython\_plot'' to visualize the explanations generated by each explainer separately, or launch a dashboard to visualize all the explanations at the same time to obtain more insights about the underlying model.
\begin{figure}[!ht]
\center
\includegraphics[width=0.8\linewidth]{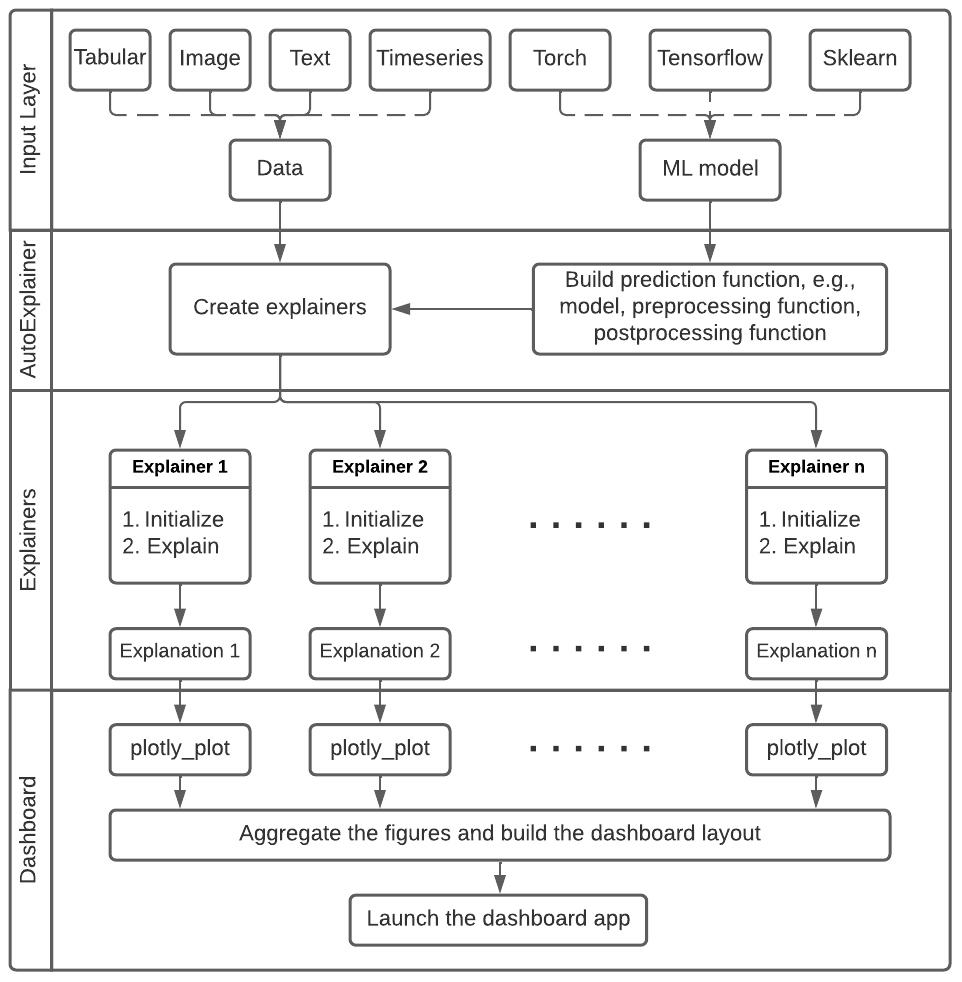}
\caption{The pipeline for generating explanations in OmniXAI}
\label{fig:pipeline}
\end{figure}
Figure \ref{fig:pipeline} shows the pipeline for generating explanations.

Suppose we have a machine learning model to explain with the training dataset ``tabular\_data'' and the feature processing function ``transformer.transform''. Here is the sample code for creating a ``TabularExplainer''. 
\begin{lstlisting}[language=Python]
from omnixai.explainers.tabular import TabularExplainer
# Initialize a TabularExplainer
explainer = TabularExplainer(
  explainers=["lime", "shap", "mace", "pdp"],  # The explainers to apply
  mode="classification",   # The task type
  data=tabular_data,       # The data for initializing the explainers
  model=model,             # The machine learning model to explain
  preprocess=preprocess,   # Converting raw features into the model inputs
  params={
     "mace": {"ignored_features": ["Sex", "Race", "Relationship", "Capital Loss"]}
  }                        # Additional parameters
)
\end{lstlisting}
In this example, LIME, SHAP, and MACE generate local explanations while PDP generates global explanations. Given the test instances, ``explainer.explain'' returns the local explanations generated by the three methods, and ``explainer.explain\_global'' returns the global explanations generated by PDP. 
\begin{lstlisting}[language=Python]
# Generate local and global explanations
local_explanations = explainer.explain(X=test_instances)
global_explanations = explainer.explain_global()
\end{lstlisting}
Finally, given the generated explanations, we provide visualization tools to visualize the results.
A dashboard can be configured by setting the test instances, the generated local explanations, the generated global explanations, the class names, and additional parameters for visualization (e.g., only plotting the selected features in PDP).
\begin{lstlisting}[language=Python]
from omnixai.visualization.dashboard import Dashboard
# Launch a dashboard for visualization
dashboard = Dashboard(
   instances=test_instances,                # The instances to explain
   local_explanations=local_explanations,   # Set the local explanations
   global_explanations=global_explanations, # Set the global explanations
   class_names=class_names,                 # Set class names
   params={"pdp": {"features": ["Age", "Education-Num", "Capital Gain", "Capital Loss", "Hours per week", "Education", "Marital Status", "Occupation"]}}
)
dashboard.show()                            # Launch the dashboard
\end{lstlisting}

\section{Supported Explanation Methods}

There are two types of explanation methods, i.e., model-agnostic and model-specific. ``Model-agnostic'' means the methods can explain the decisions made by a black-box machine learning model without knowing the model details. ``Model-specific'' means the methods require some knowledge about the model to generate explanations, e.g., whether the model is a linear model or whether the model is differentiable w.r.t. its inputs. Model-specific methods provide more explanations about the decisions, e.g., the decision paths in a decision tree model.

\subsection{Model-agnostic Explanation}

The library treats all the methods for data analysis as model-agnostic methods because they analyze training/test data directly without using trained machine learning models, e.g., feature correlation analysis. Besides data analysis, some popular model-agnostic explanation methods, e.g., LIME~\citep{Ribeiro2016}, SHAP~\citep{NIPS2017_7062}, L2X~\citep{chen2018learning}, Partial Dependence Plots (PDP)~\citep{hastie01statisticallearning}, and Morris sensitivity analysis~\citep{10.2307/1269043} are included in the library. LIME, SHAP, and L2X generate local explanations, i.e., explaining a particular decision, while PDP and sensitivity analysis generate global explanations, i.e., explaining model behavior in general.

\subsection{Model-specific Explanation}

The model-specific methods in the library include gradient-based methods, e.g., the integrated-gradient (IG) method~\citep{10.5555/3305890.3306024}, Grad-CAM~\citep{Grad-CAM2019}, etc., the contrastive explanation method (CEM)~\citep{dhurandhar2018explanations}, and feature visualization. These methods need to compute the gradients of model outputs with respect to model inputs or particular internal layers, and thus require that the models to explain are neural networks implemented by PyTorch or Tensorflow. Different from gradient-based methods, CEM solves an optimization problem to compute pertinent negatives and pertinent positives for explanation, which can utilize numerical gradients instead of analytical gradients during optimization. Because computing numerical gradients is relatively expensive with high-dimensional data, our current CEM only supports differentiable models and thus is classified as a model-specific method. Future work will make CEM support gradient estimation methods for high-dimensional data.

\subsection{Counterfactual Explanation}

Counterfactual explanation interprets a model's decision on a query instance by generating counterfactual examples that have minimal changes w.r.t the query instance’s features to yield a predefined output~\citep{Wachter2017CounterfactualEW,Moraffah2020,Byrne2019}. It can not only explain the outcome of a model's decision but also provide insight on how to change the outcome in the future. Although such methods can be grouped into ``model-agnostic'' and ``model-specific'', we highlight them by creating a separate package in the library. 

For tabular and time-series data, we recommend the recent MACE method~\citep{mace} to generate counterfactual examples. MACE is a novel framework of model-agnostic counterfactual explanation, adopting a newly designed pipeline that can efficiently handle black-box machine learning models on a large number of feature values. 


For image data, we choose the algorithm discussed in the paper~\citep{wachter2018a}. This algorithm solves the following optimization problem given a query instance $\x$:
\begin{equation}
    \min_{\x'}\max_{\lambda} \lambda\mathcal{H}(f_y(\x') - \max_{y' \neq y}f_{y'}(\x')) + \|(\x' - \x)\|_1,
\end{equation}
where $\mathcal{H}$ is the hinge loss function, $f_y(\x)$ represents the predicted class probability of sample $\x$ and label $y$, and the L1-norm regularization term is applied to encourage sparse modifications. Intuitively, it finds a counterfactual example $\x'$ with a different predicted label from $\x$ and minimum changes of $\x$. This approach can also be applied to tabular data if all the features are continuous-valued.

For text data, we utilize the pretrained model ``Polyjuice''~\citep{wu-etal-2021-polyjuice} to generate a set of potential counterfactual examples for classification tasks and question-answering tasks, and then these examples are sorted based on whether the predicted labels/results are different from the original labels/results and the distances to the original texts.

\section{Experiments}
In this section, we present four sets of experiments to demonstrate the four major types of XAI capabilities in our library to deal with four types of real data.  

\subsection{XAI for Tabular Data: Income Prediction}

The first experiment is to explain tabular data of an income prediction task. The dataset used in this example is the Adult dataset\footnote{https://archive.ics.uci.edu/ml/datasets/adult} including 14 features, e.g., age, workclass, education. The goal is to predict whether income exceeds \$50K per year on census data. We trained an XGBoost classifier~\citep{Chen:2016:XST:2939672.2939785} for this task, and then OmniXAI is applied to generate explanations for each prediction given test instances.
\begin{figure}[t] 
\center
\includegraphics[width=0.99\linewidth]{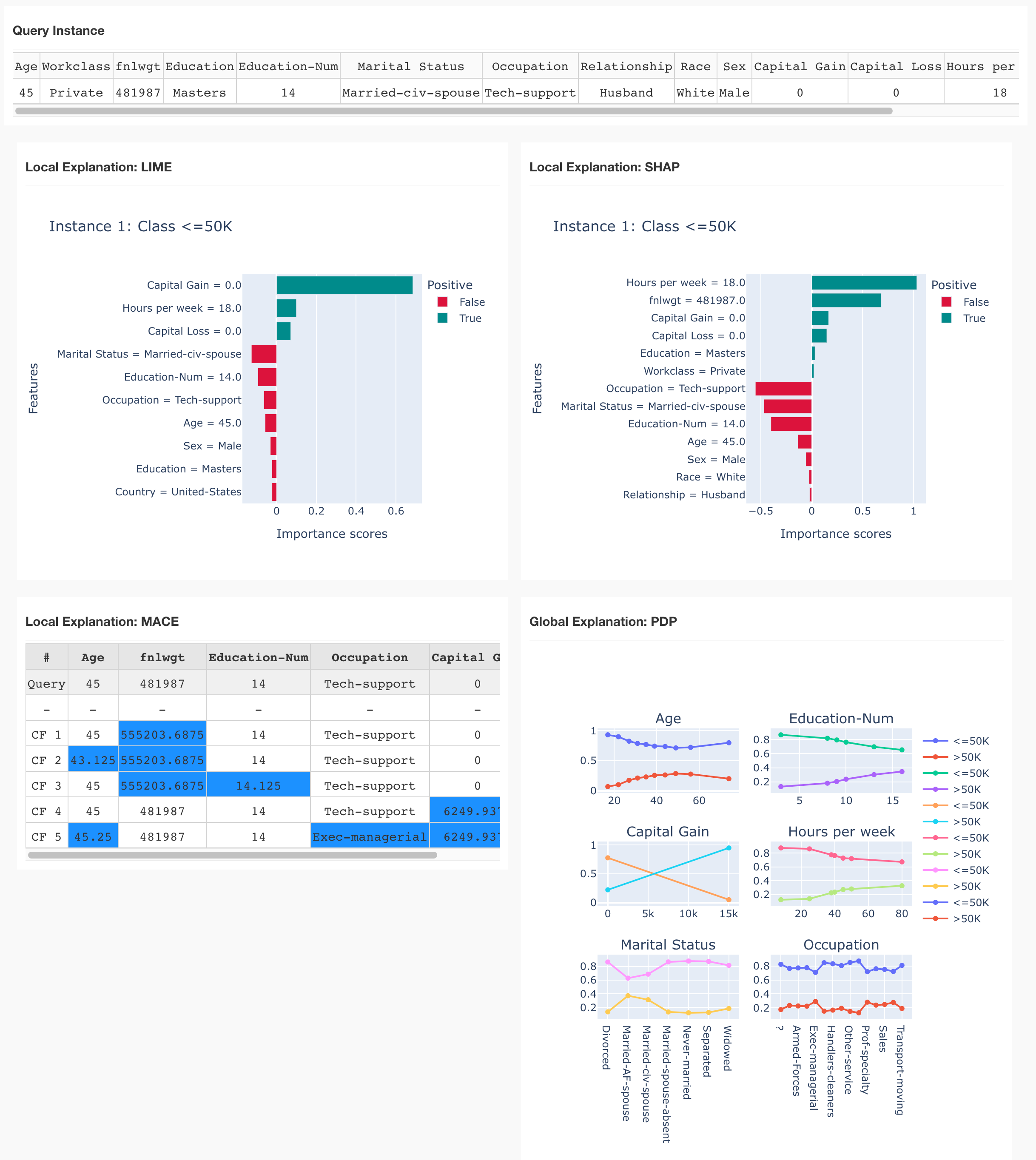}
\caption{Dashboard visualization for income prediction}
\label{fig:income}
\end{figure}

Figure \ref{fig:income} shows the explanation results generated by multiple explainers given a test instance, e.g., LIME, SHAP, MACE, and PDP. One can easily compare the feature-attribution explanations generated by LIME and SHAP, i.e., the predicted class is ``$\leq 50k$'' because ``Hours per week = 18'' (less than normal working hours), ``Capital gain = 0'' (no additional asset income), and ``Captial loss'' (people whose income is larger than 50k may have more investments). MACE generates counterfactual examples, exploring ``what-if'' scenarios and obtaining more insights of the underlying ML model, e.g.,  if ``Capital gain'' increases from 0 to 6249, the predicted class will be ``$\ge 50k$'' instead of ``$\leq 50k$''. PDP explains the overall model behavior, e.g., income will increase as ``Age'' increase from 20 to 45 while income will decrease when ``Age'' increase from 45 to 80, longer working hours per week may lead to a higher income, and married people have a higher income (this is a potential bias in the dataset). From this dashboard, one can easily understand why such a prediction is made, whether there are potential issues in the dataset, and whether to do more feature processing/selection.

\subsection{XAI for Vision: Image Classification} 

\begin{figure}[t] 
\center
\includegraphics[width=0.8\linewidth]{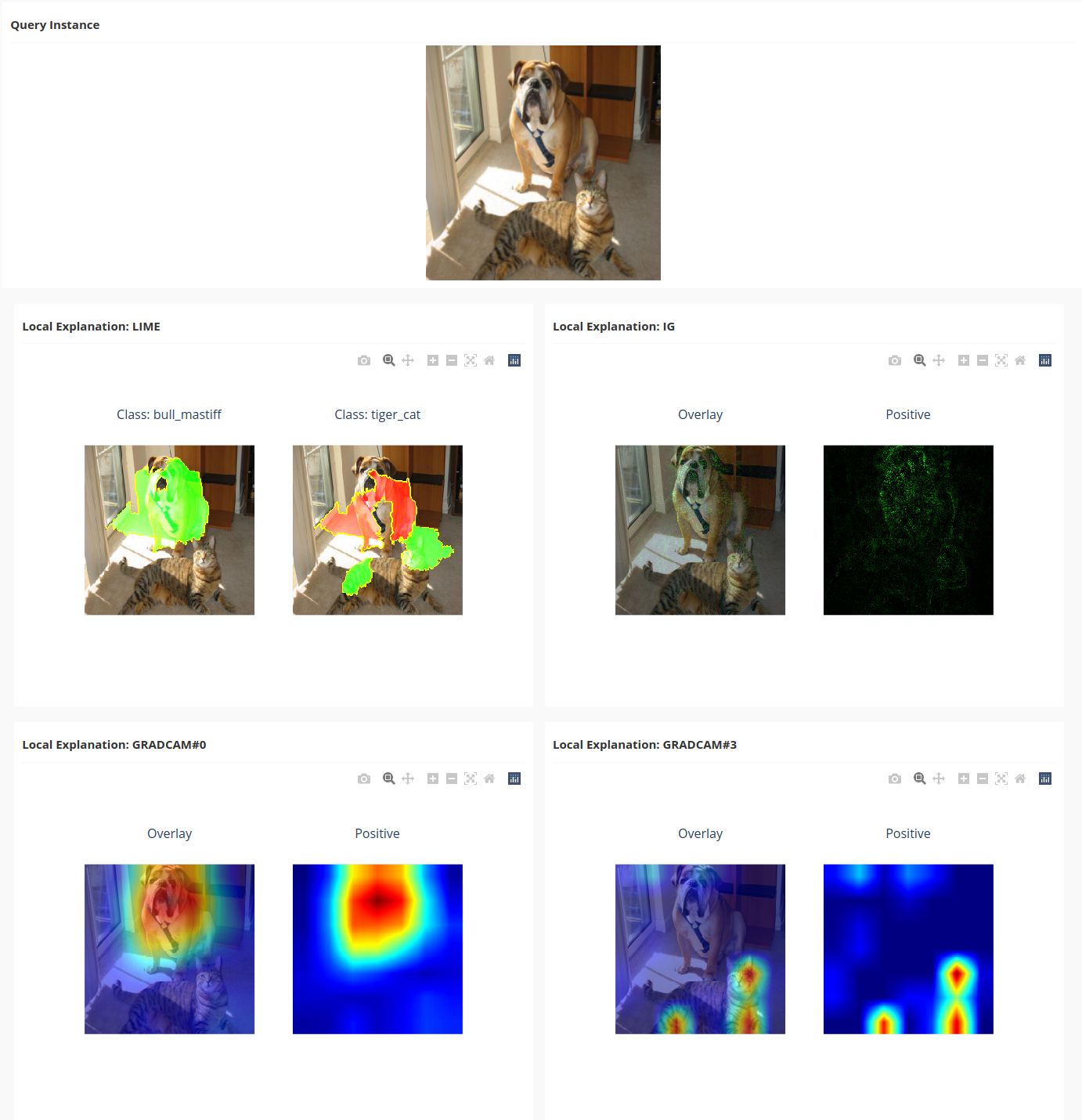}
\caption{Dashboard visualization for image classification}
\label{fig:image_classification_2}
\end{figure}

The second experiment is to explain an image classification task. We choose a ResNet~\citep{DBLP:journals/corr/HeZRS15} pretrained on ImageNet~\citep{deng2009imagenet}. Here is the sample code:
\begin{lstlisting}[language=Python]
explainer = VisionExplainer(
    explainers=["lime", "ig", 
        "gradcam#0", "gradcam#3"],
    mode="classification",
    model=model,              # A PyTorch ResNet50 model
    preprocess=preprocess,    # The preprocessing function
    postprocess=postprocess,  # The postprocessing function
    params={
        # Set the target layer
        "gradcam#0": {"target_layer": model.layer4[-1]},
        "gradcam#3": {"target_layer": model.layer4[-3]},
    },
)
local_explanations = explainer.explain(
    img,
    # Explain a different label, e.g., "y=281" corresponds to "tiger_cat"
    params={"gradcam#3": {"y": [281]}}
)
\end{lstlisting}
Figure \ref{fig:image_classification_2} shows the explanations generated by multiple explainers, e.g., Grad-CAM, Integrated-gradient, LIME.
The top predicted label of this test instance is ``bull\_mastiff''. These explainers explain the top predicted label by default (one can also set other labels to explain), e.g., integrated-gradient highlights the regions corresponding to ``bull\_mastiff''. Note that besides generating explanations with different explainers, OmniXAI can also generate explanations with the same explainer but different parameters. In this example, we apply Grad-CAM with different parameters, e.g., the target layer of ``gradcam0'' is ``layer4[-1]'' while the target layer of ``gradcam3'' is ``layer4[-3]'', and the label to explain for the first Grad-CAM explainer is ``bull\_mastiff'' (the top one label) while the label for the second Grad-CAM explainer is ``tiger\_cat''.

\begin{figure}[t] 
\center
\includegraphics[width=0.9\linewidth]{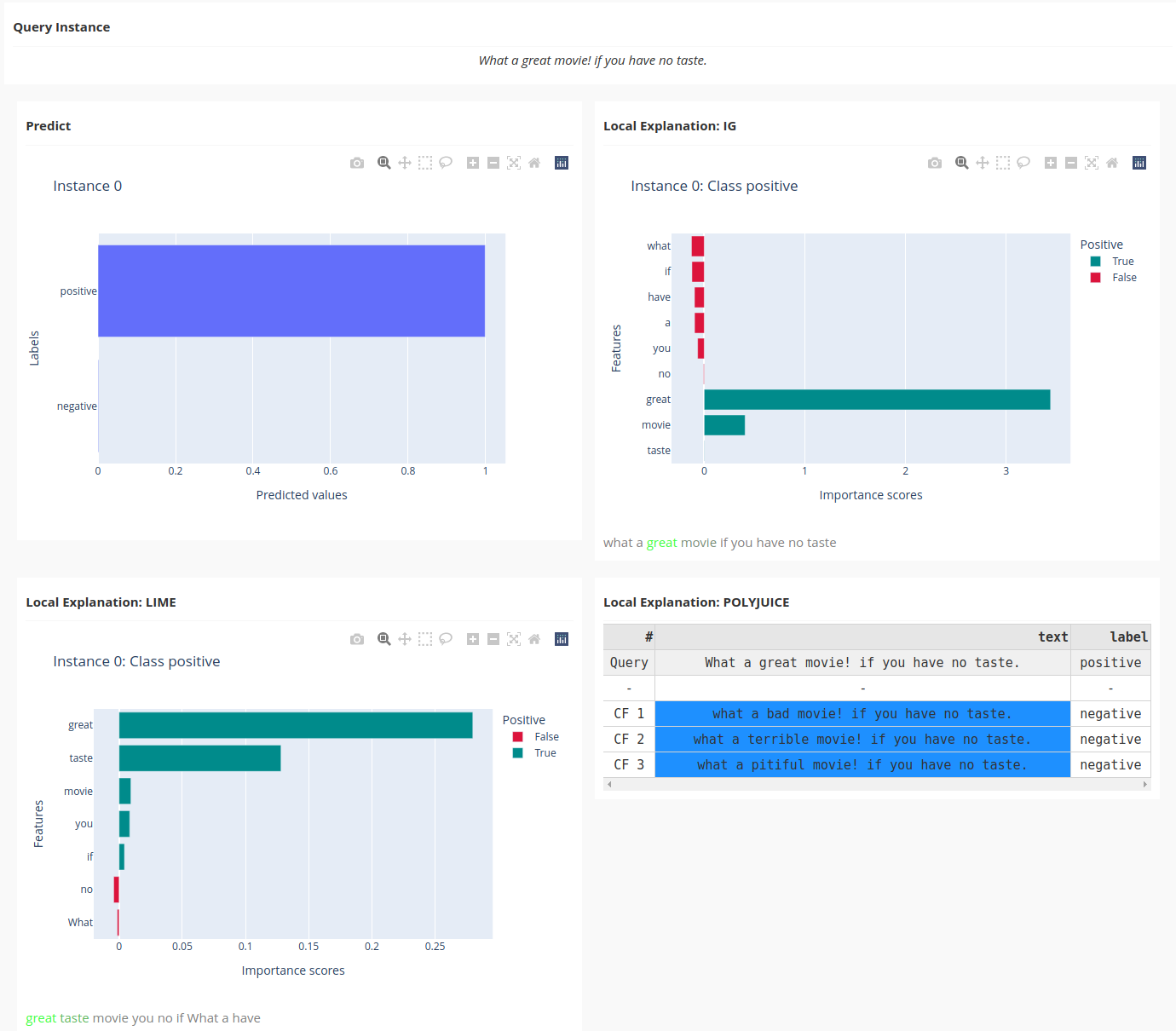}
\caption{Dashboard visualization for text classification}
\label{fig:text_classification}
\end{figure}

\subsection{XAI for NLP: Sentiment Classification} 

The third experiment is to demonstrate the explanation for NLP tasks. Specifically, we consider a sentiment classification task on the IMDB dataset where the goal is to predict whether a user review is positive or negative. We train a text CNN model for this classification task using PyTorch, and then apply OmniXAI to generate explanations for each prediction given test instances. Suppose the processing function that converts the raw texts into the inputs of the model is  "preprocess", and we want to analyze word/token importance and generate counterfactual examples. The following code shows how to do this:
\begin{lstlisting}[language=Python]
# The preprocessing function
preprocess_func = lambda x: tuple(torch.tensor(y) for y in preprocess(x))
# The postprocessing function
postprocess_func = lambda logits: torch.nn.functional.softmax(logits, dim=1)
# Initialize a NLPExplainer
explainer = NLPExplainer(
    explainers=["ig", "lime", "polyjuice"],
    mode="classification",
    model=model,
    preprocess=preprocess_func,
    postprocess=postprocess_func,
    params={"ig": 
        # The embedding layer in the model
        {"embedding_layer": model.embedding,
        # For converting token ids into tokens
         "id2token": id_to_token}}
)

x = Text("What a great movie! if you have no taste.")
# Generates explanations
local_explanations = explainer.explain(x)
# Launch a dashboard for visualization
dashboard = Dashboard(
    instances=x,
    local_explanations=local_explanations,
    class_names=class_names
)
dashboard.show()
\end{lstlisting}
Figure \ref{fig:text_classification} shows the explanation results generated by LIME, Integrated gradient (IG) and counterfactual explanation. Clearly, LIME and IG show that the word “great” has the largest word/token importance score, which implies that the sentence “What a great movie! if you have no taste.” is classified as “positive” because it contains “great”. The counterfactual method generates several counterfactual examples for this test sentence, e.g., “what a terrible movie! if you have no taste.”, helping us understand more about the model behavior.

\begin{figure}[t] 
\center
\includegraphics[width=0.9\linewidth]{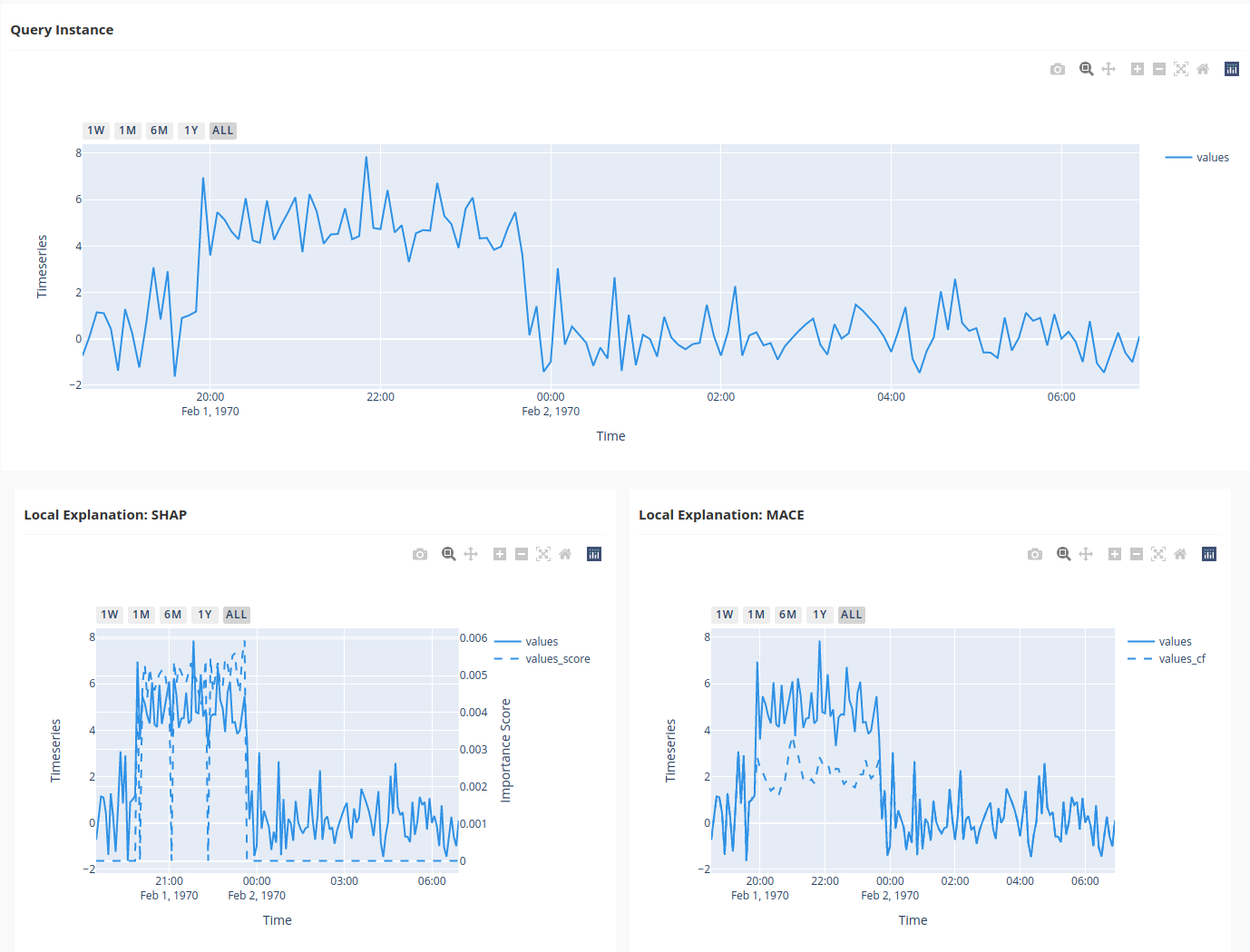}
\caption{Dashboard visualization for time-series anomaly detection.}
\label{fig:timeseries}
\end{figure}

\subsection{XAI for Time-Series: Anomaly Detection} 

The last experiment is to explain time-series data. We consider a univariate time-series anomaly detection task. We use a simple statistics-based detector for demonstration, e.g., a window of time-series is detected as an anomaly according to some threshold estimated from the training data. Suppose we have detector ``detector'', training data ``train\_df'' and a test instance ``test\_df''. The following code shows how to apply OmniXAI in anomaly detection:
\begin{lstlisting}[language=Python]
# Initialize a TimeseriesExplainer
explainers = TimeseriesExplainer(
    explainers=["shap", "mace"],        # Apply SHAP and MACE explainers
    mode="anomaly_detection",           # An anomaly detection task
    data=Timeseries.from_pd(train_df),  # Set data for initializing the explainers
    model=detector,                     # Set the black-box anomaly detector
    preprocess=None,
    postprocess=None,
    params={"mace": {"threshold": 0.001}}   # Additional parameters for MACE
)
# Generate local explanations
test_instance = Timeseries.from_pd(test_df)
local_explanations = explainers.explain(test_instance)
# Launch a dashboard for visualization
dashboard = Dashboard(instances=test_instance, local_explanations=local_explanations)
dashboard.show()
\end{lstlisting}

Figure \ref{fig:timeseries} shows the explanation results generated by SHAP and MACE. The dash lines demonstrate the importance scores and the counterfactual examples, respectively. SHAP shows the most important timestamps that make this test instance detected as an anomaly. MACE provides a counterfactual example showing that it will not be detected as an anomaly if the metric values from 20:00 to 00:00 are around 2.0. From these two explanations, one can clearly understand the reason why the model considers it as an anomaly.

\section{Broader Impacts and Responsible Use}

The adoption of OmniXAI in real-world applications potentially can yield a broad range of positive impacts. OmniXAI offers a set of comprehensive and effective tools to interpret AI models and explain their decisions to improve the transparency and trustworthiness of AI systems, thus helping developers and users understand the logic behind the decisions. AI models sometimes may make unsatisfying or wrong decisions in real-world applications due to the data shift, sampling, labeling bias or other real data complexities. Explaining why an AI model fails is thus important and necessary, which is the key to keeping users confident in AI systems and help developers improve model performance. OmniXAI can be used to analyze various aspects of AI models and help developers figure out the issues and causes when a wrong decision occurs so that developers can quickly understand the reasons behind failures and improve the models accordingly. We encourage researchers, data scientists, and ML practitioners to adopt OmniXAI in real-world applications for positive impacts, e.g., improving transparency, accuracy, bias and fairness of AI/ML models.

While OmniXAI has many potential positive impacts, the misuse of OmniXAI may result in negative impacts. We encourage users and readers to read the detailed discussion and the guidelines for the responsible use of explainable ML in \citep{Patrick2019}. 
In particular, the OmniXAI library should not be used to explain misused AI/ML models or any unethical use case. It also should not be used to enable hacking of models through public prediction API, or stealing black-box models or breaking privacy policies~\citep{Rudin2018,Florian2016,Reza2019}.
For practical uses in real-world applications, we recommend using multiple explainers supported in OmniXAI, e.g., SHAP for feature-attribution explanation, MACE for counterfactual explanation, PDP for global explanation or ``glass'' models such as linear or tree-based models, for explaining a particular model decision to reduce the risk of adversarial manipulation. Finally, we note that while OmniXAI can help to understand and improve the sociological bias when used properly, it is not guaranteed to detect all kinds of sociological biases, which could be further improved in future work. 

\section{Conclusions and Future Work}

We present OmniXAI, a comprehensive open-source library for explainable AI, which provides omni-way explanation capabilities to address many of the pain points in understanding and interpreting the decisions made by machine learning models. It supports various data types and tasks, and provides unified, easily extensible interfaces for a wide range of explanation methods, e.g., feature-attribution explanation, counterfactual explanation, gradient-based explanation, etc. It also includes a GUI dashboard for users to examine explanation results and compare different interpretable algorithms. We will continue our efforts to improve OmniXAI. Some planned future work includes adding more techniques for data analysis (e.g., detecting mislabeled samples and bias issues), adding more latest interpretable ML algorithms especially for NLP and time-series, and adding supports for other types of data and tasks. We will also improve the dashboard and user interface to make it easier to use and more comprehensive. Finally, we welcome any feedback and suggestions for improvement, and encourage contributions from the open-source communities. 

\section*{Acknowledgements}

We would like to thank our colleagues and leadership teams from Salesforce who have provided strong support, advice, and contributions to this project. We also want to thank our ethical AI team especially for Anna Bethke for the valuable feedback on this work.

\bibliography{ref}

\end{document}